\documentclass[letterpaper, 10 pt, conference]{ieeeconf}  
\IEEEoverridecommandlockouts                              

\usepackage{graphics}
\usepackage{float} 
\usepackage{caption}
\usepackage{subfigure}
\usepackage{epsfig}
\usepackage{mathptmx}
\usepackage{times}
\usepackage{amsmath} 
\usepackage{amssymb}  
\usepackage{array}
\usepackage{url} 
\usepackage{color}
\usepackage{cite} 
\usepackage[ruled,linesnumbered,noend]{algorithm2e}  
\usepackage{siunitx}
\usepackage{alphalph}
\title{\LARGE \bf
	Ada-Detector: Adaptive Frontier Detector for Rapid Exploration
}

\author{
	Zezhou Sun$^{\dag}$,
	Banghe Wu$^{\dag}$,
	Chengzhong Xu$^{\ddag}$,
	Hui Kong$^{*}$
	\thanks{$\dag$ School of Computer Science and Engineering, Nanjing University of Science and Technology, Nanjing, Jiangsu, China.}
	\thanks{$\ddag$ The State Key Laboratory of Internet of Things for Smart City (SKL-IOTSC), Department of Computer Science, University of Macau, Macau, China.}
	\thanks{$*$ The State Key Laboratory of Internet of Things for Smart City (SKL-IOTSC), Department of Electromechanical Engineering (EME), University of Macau, Macau, China.}
	\thanks{
	This work is supported by National Key Research and Development Program of China (No. 2019YFB2102100), the Science and Technology Development Fund of Macau SAR (File no. 0015/2019/AKP), Guangdong-Hong Kong-Macao Joint Laboratory of Human-Machine Intelligence-Synergy Systems (No. 2019B121205007), and the National Natural Science Foundation of China (No.61803083). }
}

\begin{document}
	
	\maketitle
	\thispagestyle{empty}
	\pagestyle{empty}
	
	\begin{abstract} 
	
	In this paper, we propose an efficient frontier detector  method based on adaptive Rapidly-exploring Random Tree (RRT) for autonomous robot exploration. Robots can achieve real-time incremental frontier detection when they are exploring unknown environments. 
	First, our detector adaptively adjusts the sampling space of RRT by sensing the surrounding environment structure. The adaptive sampling space can greatly improve the successful sampling rate of RRT (the ratio of the number of samples successfully added to the RRT tree to the number of sampling attempts) according to the environment structure and control the expansion bias of the RRT. 
    Second, by generating non-uniform distributed samples, our method also solves the over-sampling problem of RRT in the sliding windows, where uniform random sampling causes over-sampling in the overlap area between two adjacent sliding windows. In this way, our detector is more inclined to sample in the latest explored area, which improves the efficiency of frontier detection and achieves incremental detection. 
    We validated our method in three simulated benchmark scenarios. The experimental comparison shows that we reduce the frontier detection runtime by about $40\%$ compared with the SOTA method, DSV Planner. 
    
	\end{abstract}
	
	\section{INTRODUCTION}
	
	The frontier detection module is a key module for mobile robots to autonomously explore unknown environments. 
	Frontiers are the boundaries separating known space from unknown area. 
    The frontier detection algorithms detect the frontiers according to the map established in real-time during the movement of the robot and send the frontier with the highest priority to the path planning module. 
    The ideal frontier detection module deployed to an autonomous exploratory robot is expected to detect new frontiers within the interval between two LiDAR scans and empowers robots to replace humans to achieve efficient and robust autonomous exploration in various environments. 
    
    Since the RRT algorithm \cite{lavalle1998rapidly} is heavily biased to grow towards unknown regions of the map \cite{umari2017autonomous} and can easily be extended to high-dimensional spaces, it is widely used for frontier detection tasks in the field of autonomous exploration.
    The original RRT-based frontier detector \cite{bircher2016receding} extends an RRT in the free space (the space that has been observed by sensors and is not occupied by obstacles) and treats the nodes in the RRT as viewpoints. The exploration gain of each viewpoint is calculated according to the sensor coverage. The viewpoints whose exploration gain is greater than the preset threshold are called frontiers. Thereafter, the robot sorts the frontiers according to the exploration gain and drives to the frontier with the highest priority. 

	Compared with path planning, since frontier detection does not have a clear expansion target, it can only use the RRT method to expand to the surrounding area uniformly instead of using RRT improvement methods such as the Informed RRT* \cite{gammell2014informed}, Batch Informed trees (BIT*) \cite{gammell2015batch}, and Regionally accelerated batch informed trees (RABIT*) \cite{choudhury2016regionally} etc. 
	Therefore, the problem of how to improve the efficiency of frontier detection has attracted the attention of researchers. The Graph-Based subterranean exploration path Planner (GBPlanner) \cite{dharmadhikari2020motion} uses a periodic sliding window to restrict the sampling space of RRT to the surrounding region of the robot. 
	Compared to covering the entire exploration area, RRT can fill the sliding window more quickly. 
	Thus, this method significantly improves the efficiency of frontier detection. 
	
	\begin{figure}
		\centering
		\subfigure[]{
			\includegraphics[width=0.28\linewidth]{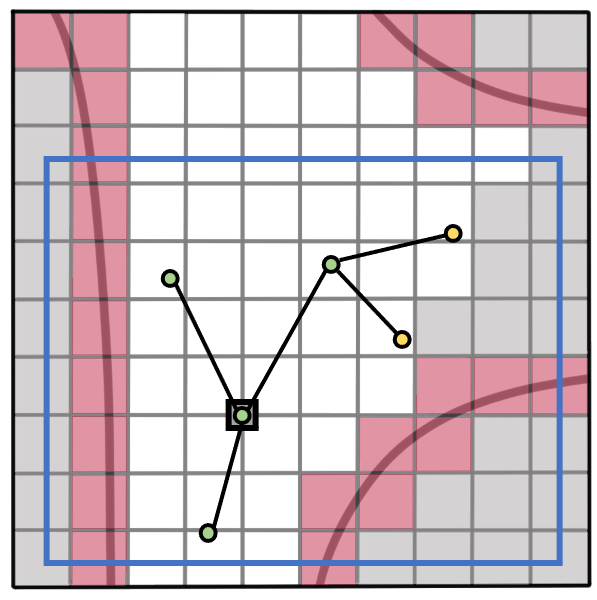}
		}
		\subfigure[]{
			\includegraphics[width=0.28\linewidth]{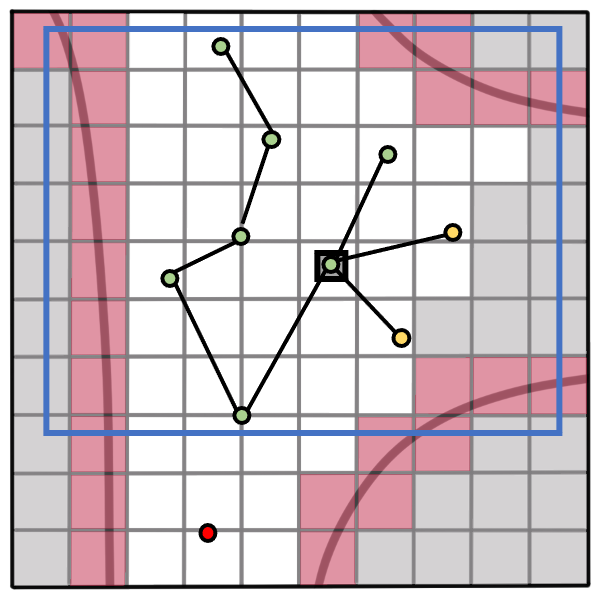}
		}
		\subfigure[]{
			\includegraphics[width=0.28\linewidth]{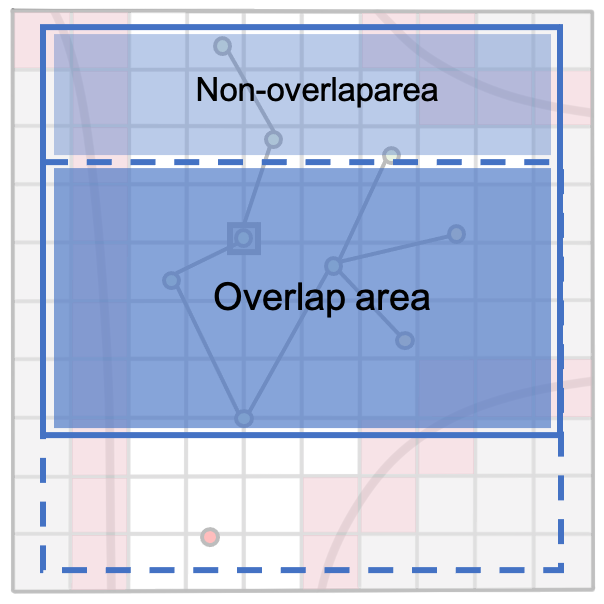}
		}
		\subfigure{
			\includegraphics[width=1\linewidth]{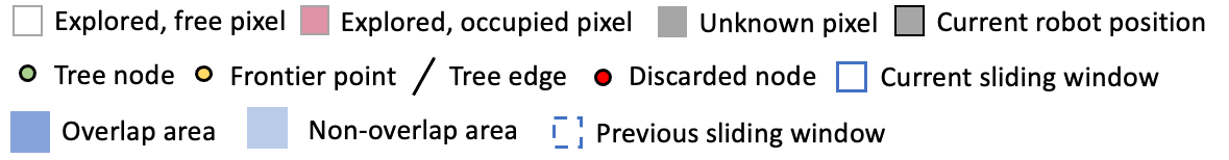}
		}
		\caption{
		An illustration of the dynamically expanded frontier detector in DSV Planner \cite{dsvp}. 
        (a) shows the RRT-based frontier detection performed in a sliding window. 
        (b) After the sliding window is updated with the movement of the robot, DSV Planner \cite{dsvp} discards the node that is beyond the updated sliding window and continues expanding the RRT tree. 
        (c) In two adjacent sliding windows, the overlap area is repeatedly sampled twice and the non-overlap area is the new observation area. 
        }
		\label{dsvp}
	\end{figure}
	
	Usually, the size of the sliding window is set as the sensing range limit of the sensor on the robot.
	However, with the improvement of sensor manufacturing technology, the sensing range limit of a modern multi-channel LiDAR has exceeded 100m. Thus, the sliding window bounded by the sensing range limit is too large for real-time frontier detection. 
	In addition, the use of a fixed-sized sliding window cannot fully reflect the robot's perception of the surrounding structure of the environment. For example, when passing through a narrow corridor, it is preferred that the RRT should expand in the direction of the corridor. When passing an intersection, it should pay more attention to the left and right sides. When entering a room, the sampling space of RRT should be limited by walls. In contrast, the fixed-sized sliding window enables the RRT to expand uniformly in all directions. 
	Through the sensor's perception of the surrounding environment structure, we can adaptively reduce the size of the sliding window to the smallest circumscribed rectangle of LiDAR scan.
	
	Furthermore, there is inevitably an overlap area between two adjacent sliding windows. 
	The non-overlap area of the current sliding window actually corresponds to the new observed (explored) area of the LiDAR sensor. 
	The GBPlanner \cite{dharmadhikari2020motion} uniformly generates sampling points in each sliding window.
	Thus, the overlap areas are twice as dense as the non-overlap area due to repeated sampling. We call this phenomenon an over-sampling problem, as shown in Fig. \ref{dsvp}(c). 
	However, it is preferred that the detector should focus more on detecting frontiers in the new observed area. 
	Obviously, the over-sampling problem can reduce the tendency of extending the RRT tree to a new observed area. 
	
    Very recently, the Dual-Stage Viewpoint Planner (DSV Planner) \cite{dsvp} proposed a dynamically extended frontier detector. 
    Figure \ref{dsvp} shows the dynamic expansion process. 
    In Fig. \ref{dsvp}(a), the blue box is the sliding window,  representing the sampling space of RRT. It expands an RRT tree to detect the frontiers in the sliding window. 
    When the robot moves to the new position in Fig. \ref{dsvp}(b), the sliding window is moved accordingly to the new position of the robot (the blue box in (b)). 
    Nodes in the RRT tree that is beyond the sliding window are discarded. 
    Iteratively, the robot continues expanding the RRT tree in the updated sliding window. 
    Once detecting the frontiers, DSV Planner \cite{dsvp} selects the frontiers with the highest priority and additionally increases the sampling probability within the neighborhood of these frontiers. 
    When the number of nodes in the RRT tree exceeds a certain threshold or the number of sampling attempts reaches the upper limit, the robot considers that the frontier detection in this sliding window is completed. 
    Then the detector sends the frontier with the highest priority to the path planning module. 
    
    However, the DSV Planner \cite{dsvp} still makes lots of sampling attempts in the overlap area, which is hardly helpful for the robot to travel to the newly explored frontiers. 
	In addition, this method only increases the bias of the RRT expansion toward some specific frontiers, causing the detector to pay insufficient attention to the newly explored area that is beyond the neighborhood of selected frontiers.
	It can also make RRT trapped into obstacle regions in a complex environment. To deal with these issues, we counteract over-sampling and achieve incremental detection by generating non-uniform samples.
    
    \begin{figure}
		\centering
		\subfigure[]{
			\includegraphics[width=0.42\linewidth]{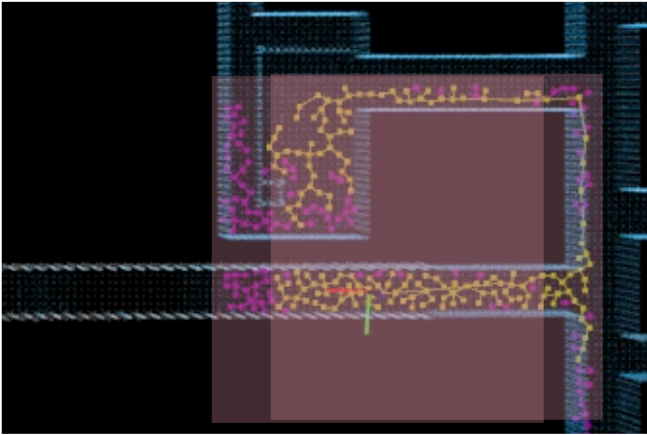}
		}
		\subfigure[]{
			\includegraphics[width=0.42\linewidth]{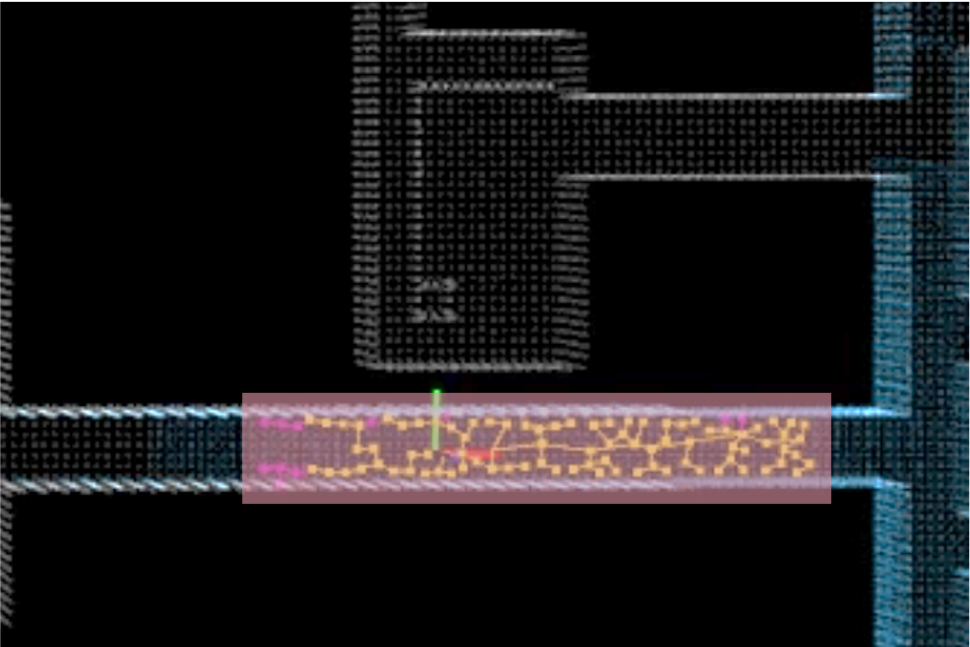}
		}
		\subfigure{
			\includegraphics[width=1\linewidth]{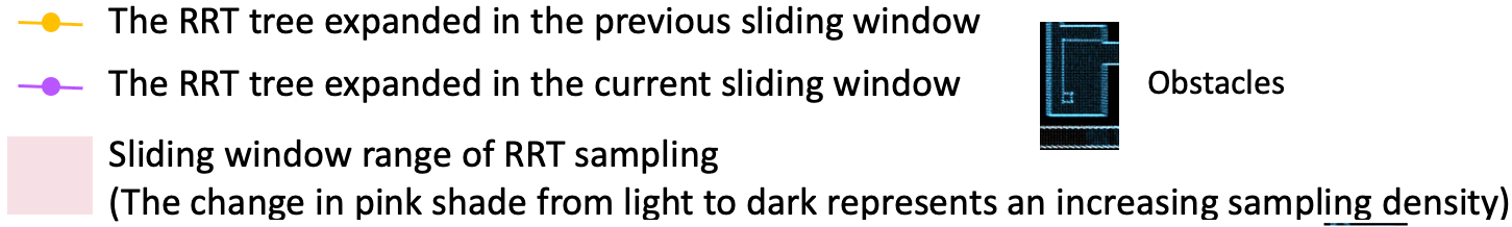}
		}
		\caption{
		An illustration of the improvement of our method from the previous method. 
		Compared with the previous methods that detect frontiers in fixed-sized sliding windows, our method adaptively adjusts the sliding windows through the perception of the surrounding structural environment. 
		Our detector also generates non-uniform distributed samples to solve the over-sampling problem in the overlap area between two adjacent sliding windows. 
        }
		\label{firstpic}
	\end{figure}
    
	

	The main contributions of this paper are as follows. (1) We propose an adaptive frontier detector with a varying-sized sliding window, which can improve the efficiency of frontier detection by limiting the sampling space of RRT. 
	(2) We generate non-uniform samples to avoid over-sampling between two adjacent sliding windows, making RRT biased to detect frontiers in newly explored areas and achieve incremental detection.

	The remainder of this paper is organized as follows.
    Section II outlines the related works.
    The details of our method are given in Section III.
    Experimental results are shown in Section IV, followed by conclusions in Section V.
	
	\section{Related works}

	Frontier detection divides the created occupancy map into free area, unknown area, and occupied area according to the confidence and searches for the boundaries between free area and unknown area in the map \cite{yamauchi1997frontier}. 
	According to two different path planning algorithms, Dijkstra \cite{frana2010interview} and RRT, frontier detection is mainly divided into two categories: search-based and sampling-based. 
	
    \subsubsection{The search-based frontier detectors}
	Yamauchi et al. \cite{yamauchi1997frontier} proposed their seminal search-based exploration work based on frontier detection and tracking.
	In order to detect frontiers more efficiently, the following-up works are devoted to reducing the search space of frontiers. 
	The Wavefront Frontier Detection (WFD) \cite{keidar2012robot} reduces the search space from the entire map to the known area. 
	The Expanding-Wavefront Frontier Detection (EWFD) \cite{quin2014expanding} only incrementally searches for the latest explored areas. 
	The Dense Frontier Detection method (DFD) \cite{orvsulic2019efficient} performs frontier detection in the submaps of Cartographer \cite{hess2016real}, reducing the search space to the total frontier length of all previous submaps, 
	so that the detected frontiers are not affected by the graph optimization to the map. 
	The improved DFD method based on the sliding window \cite{DBLP:conf/iros/SunW0SYK20} reduce the search space to the frontier pool of the latest modified submaps. 
	These methods are designed to detect all frontiers in the exploration area, however, they are mostly used for 2D exploration and are difficult to extend to high-dimensional spaces. 
	
	
	\subsubsection{The sampling-based frontier detectors}

	The sampling-based frontier detection methods are more suitable for robots to explore in the 3D environment. 
	Since the ``next best view" method was proposed \cite{connolly1985determination},  
	Oriolo \cite{oriolo2004srt} and Freda \cite{freda2005frontier} introduced a probabilistic planning method called Sensor-based Random Tree (SRT), a variant of RRT, to effectively sample the map. 
	The Multiple Rapidly-exploring Randomized Trees (multi-layer RRT)\cite{umari2017autonomous} improves the efficiency of frontier detection by using global and local RRT trees. 
	With more applications of unmanned aerial vehicles (UAV), RRT-based frontier detection methods have been widely used in UAV exploration, such as the Receding Horizon Next-Best-View method (NBVP) \cite{bircher2016receding} and its variants \cite{dang2018visual} \cite{papachristos2017uncertainty}. 
	However, the sampling space of these methods is the explored map that expands with the progress of robot exploration, which means that the efficiency of frontier detection continues to decrease. 
	
	Recently, the Graph-Based exploration path Planner (GBPlanner) \cite{dharmadhikari2020motion} and Motion-primitive Based exploration path Planner (MBPlanner) \cite{dang2020graph} implement a fast frontier detector based on periodic sliding windows. 
	It limits the sampling space from the free space of all explored areas to that of the local map around the robot within the LiDAR sensing range. 
	However, GBPlanner‘s detector \cite{dharmadhikari2020motion}  cannot instantly detect all frontiers within the LiDAR sensing range during the interval of the periodic sliding window.
    Therefore, robots can miss partial frontiers and leave some unexplored areas to be ignored. 
	
	On this basis, the Dual-Stage Viewpoint Planner (DSV Planner) \cite{dsvp} proposed an efficient exploration algorithm based on a dynamically extended frontier detector. 
	DSV Planner\cite{dsvp} achieves SOTA frontier detection efficiency by reusing part of the nodes in the previous sliding window and additionally expanding toward specific frontiers.
	
    For a fair comparison of different algorithms, Zhang et al. \cite{cmuexloration}  established a public simulation environment to evaluate exploration methods and provided the best human practice results \cite{cmuexloration}. 
	The simulation environment provides 5 different scenarios including multi-layer garage, indoor, forest, tunnel, and campus environments. Zhang et al. tested their DSV Planner \cite{dsvp} and TARE \cite{tare} with NBVP \cite{bircher2016receding}, GBPlanner \cite{dharmadhikari2020motion}, MBPlanner \cite{dang2020graph}, and the best human practice results \cite{cmuexloration} in these simulation environments.

	\section{Methodology}

	\begin{figure}[tbp]
		\centering
		\includegraphics[width=0.8\linewidth]{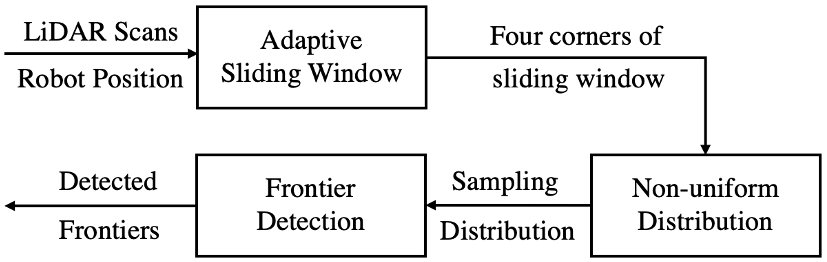}
		\caption{The system diagram of our Ada-detector. }
		\label{our-graph}
	\end{figure}

	\subsection{Adaptive Sliding Window}
	
	In the simulation environment, since it is necessary to detect slopes so that the robot can explore in 3D space, the maximum range of LiDAR sensing that can be used for mapping is set to $15$m. 
	Therefore, as did in DSV Planner \cite{dsvp},
	we do not consider the LiDAR points beyond this limit. 
	Specifically, we first align the filtered LiDAR scan with the map coordinates. 
	Then, by traversing the LiDAR points, we calculate the four corners of the smallest circumscribed rectangle of the LiDAR scan. 
	The rectangle formed by the corners is the current sliding window and the RRT regards it as the sampling space. We update the sliding window by updating the corners. 
	
\begin{figure}
		\centering
		\subfigure[]{
			\includegraphics[width=0.3\linewidth]{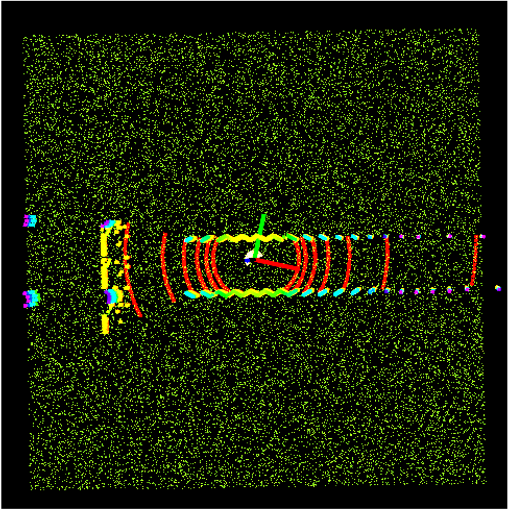}
		}
		\subfigure[]{
			\includegraphics[width=0.3\linewidth]{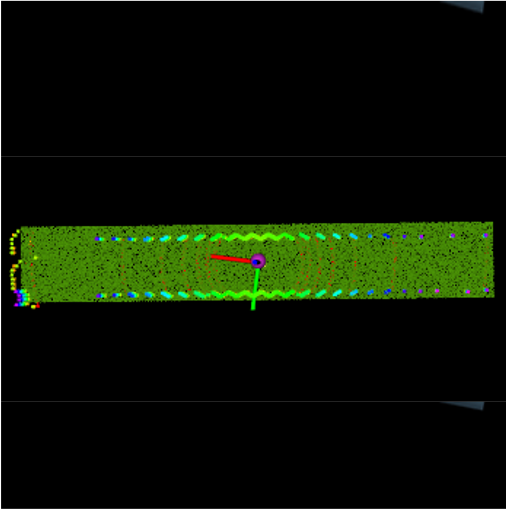}
		}
		\caption{
		(a) shows the sampling space of a fixed-sized sliding window and (b) shows the sampling space of an adaptive sliding window of our method. 
		With the same number of sampling attempts (green points), our method has a higher successful sampling rate. 
        }
		\label{adaptive}
	\end{figure}

    Generally, the successful sampling rate (the ratio of the sampled points that can be successfully added to the RRT tree to the number of sampling attempts) is directly related to the proportion of free space to the total sampling space.
    It can be seen from Fig. \ref{adaptive} that the successful sampling rate is generally much higher in an adaptive sliding window than in a fixed-sized one. 
	
	\subsection{Non-uniform Sampling within an Adaptive Sliding Window}
	
	\begin{figure}
		\centering
		\subfigure[]{
			\includegraphics[width=0.25\linewidth]{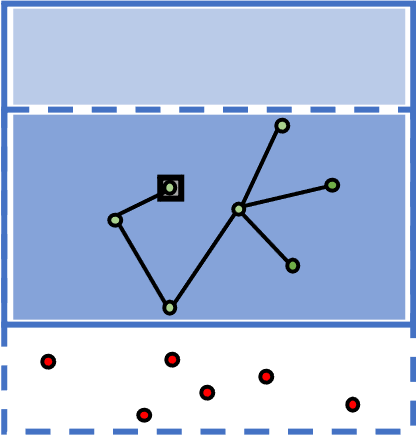}
		}
		\subfigure[]{
			\includegraphics[width=0.25\linewidth]{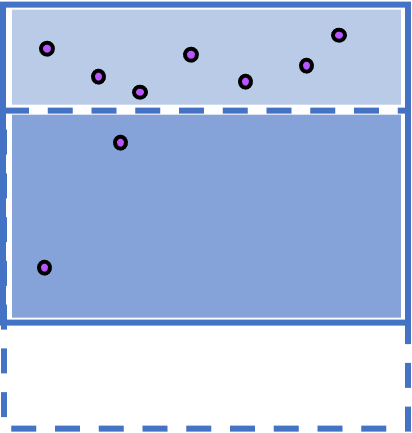}
		}
		\subfigure[]{
			\includegraphics[width=0.25\linewidth]{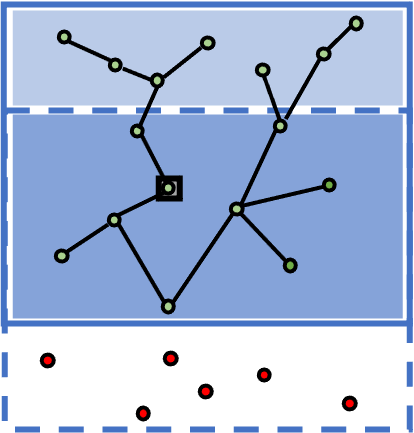}
		}
		\subfigure{
			\includegraphics[width=1\linewidth]{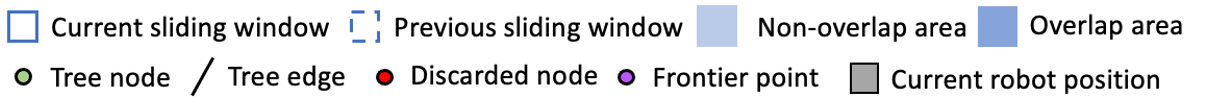}
		}
		\caption{
		An illustration of non-uniform sampling process. 
		(a) Just after the sliding window is updated, the nodes that exceed the sliding window in the RRT tree are discarded and the remaining $N_{o}$ nodes are all located in the overlap area. 
		(b) According to Alg. \ref{algorithm2}, we calculate how many points are needed to be sampled in the overlap area and the non-overlap area, respectively, to make the density of RRT nodes in the two areas equal and both larger than $\theta$. We then perform a non-uniform sampling based on this. 
		(c) We expand the RRT tree based on the non-uniform sampling. 
        }
		\label{nonuniform}
	\end{figure}
	
    To deal with the over-sampling problem of DSV Planner \cite{dsvp} as mentioned before, we give the non-uniform sampling strategy. In each new adaptive sliding window,  
    let $\theta$ be the ratio of the number of successful samples (samples successfully added to the RRT tree) to the size of the sampling space, called the successful sampling density. 
    Let $N_{total}$ be the number of RRT tree nodes, $S_{o}$ and $S_{n}$ be the area of the overlap and non-overlap region, respectively, $N_{o}$ and $N_{n}$ be the number of RRT nodes in the overlap and non-overlap region, respectively. 
    The motivation of the proposed non-uniform sampling is that, for an efficient detector, the value of $\theta$ should be the same no matter in the non-overlap or overlap area in each adaptive sliding window, i.e., $\frac{N_{o}}{S_{o}} = \frac{N_{n}}{S_{n}} = \theta$. 
    Let $\tau$ be the maximal number of successful samples in the whole adaptive sliding window, and thus we have $\theta=\frac{\tau}{S_{total}}$, where $S_{total}$ is the area of the sliding window.
    Therefore, once the condition $\theta=\frac{\tau}{S_{total}}=\frac{N_{o}}{S_{o}}= \frac{N_{n}}{S_{n}}$ is satisfied, it is guaranteed to solve the over-sampling problem in the DSV Planner \cite{dsvp}. 
    

    Note that $\tau$ is a constant for each adaptive sliding window, but $S_{total}$, $S_{o}$ and $S_{n}$ are not, meaning that $N_{o}$ and $N_{n}$ are both variables. 
    Generally, $S_{o}$ and $S_{n}$ are different. Therefore, $N_{o}$ and $N_{n}$ are not equal as well. This explains why 
    we need a non-uniform sampling within each new adaptive sliding window, where we treat the sampling in the overlap and non-overlap region differently. 
    
    Specifically,  given the area of a new updated sliding window $S_{total}$, we can obtain $\theta$. We can get the overlap area $S_{o}$ and non-overlap area $S_{n}$ based on the four corners of the previous and the current sliding window, respectively, (lines 2-3 of Alg. \ref{algorithm2}). 
    In turn, we can get the number of RRT nodes that should be successfully sampled from the overlap area, $N_{eo}$,
    and from the non-overlap area, $N_{en}$, respectively, (lines 4 of Alg.\ref{algorithm2}), as shown in Fig. \ref{nonuniform}. 
    Therefore, we can get the sampling probability $\eta_{n}$ of non-overlap area and $\eta_{o}$ of overlap area in the current sliding window to be $\frac{N_{en}}{N_{eo} + N_{en}}$ and $\frac{N_{eo}}{N_{eo} + N_{en}}$ (line 4-5 of Alg.\ref{algorithm2}), respectively .
    Once again, when the number of RRT nodes in the current sliding window equals $\tau$ or the successful sampling density is larger than $\theta$, we update the sliding window. Meanwhile, we discard the nodes in the RRT tree that are beyond the new sliding window and reserve the ones in the overlap area (line 1 of Alg.\ref{algorithm2}).

    \begin{algorithm}[htbp]
		\caption{Non-uniform sampling within an adaptive sliding window}
		\label{algorithm2}
		\KwIn{ \\
		    \quad corners of the previous sliding window $Pre_{corners}$\\ 
			\quad corners of the new sliding window $P_{corners}$\\ 
			\quad list of RRT tree nodes $RRT_{nodes}$\\
			\quad expected sampling density $\theta$ \\}
		\KwOut{ \\
			\quad sampling probability in the non-overlap area $\eta_{n}$\\
			\quad sampling probability in the overlap area $\eta_{o}$}
		
		($RRT_{nodes}$, $N_{o}$) $\leftarrow$ Discard($RRT_{nodes}$,$P_{corners}$) \\
		($S_{o}$, $S_{n}$) $\leftarrow$ overlapArea($Pres_{corners}$, $P_{corners}$) \\
		$Pre_{corners}$ = $P_{corners}$ \\
		$N_{eo}$ = $\theta * S_{o} - N_{o}$; \quad$N_{en}$ = $\theta * S_{n}$ \\
		$\eta_{n} = \frac{N_{en}}{N_{eo} + N_{en}}$; \quad $\eta_{o} = \frac{N_{eo}}{N_{eo} + N_{en}}$ \\
	\end{algorithm}
	\vspace{-15pt}
	
	\subsection{Frontier Detection}
	Given $\eta_{n}$ and $\eta_{o}$, we expand RRT in the current adaptive sliding window in a non-uniform sampling mode (Alg.\ref{algorithm3}). 
    We randomly generate a number $r$, with $r\in[0,1]$. If $r>\eta_{n}$,  we sample in the overlap area. Instead, we sample in the non-overlap area (lines 3-6). 
    After obtaining the sample $P_{rand}$, we expand the RRT tree. Lines 7-11 of Alg. \ref{algorithm3} are the standard RRT process \cite{lavalle1998rapidly}. 
    
    If the sample can be successfully added to the RRT tree, we calculate the exploration gain of this node and increase the number of successful samples by one (line 12-14).  
    The method of calculating exploration gain is the same as the one used in DSV Planner \cite{dsvp}. 
    Since we want to keep the successful sampling density greater than or equal to $\theta$ in each adaptive sliding window. Once an adaptive sliding window is updated, the successful sampling density of the overlap area of the new adaptive sliding window is approximately equal to $\theta$. From this point of view, our method is about equivalent to an incremental frontier detection.

    \begin{algorithm}[htbp]
		\caption{Frontier Detection}
		\label{algorithm3}
		\KwIn{ \\
			\quad probability of sampling in non-overlap area $\eta_{n}$ \\
			\quad list of RRT tree nodes after discarding $RRT_{nodes}$ \\
			\quad the adaptive sliding window area $S_{total}$ \\
			\quad number of RRT tree nodes after discarding $N_{o}$ \\
			\quad occupancy map $M$ \\
			\quad min exploration gain $MinExplorationGain$ \\
			\quad expected sampling density $\theta$ \\
			\quad the upper limit of the RRT nodes in each sliding window $\tau$}
		\KwOut{ \\
			\quad detected frontiers $P_{frontier}$}
		$N_{total}$ = 0 \\
		\While{$N_{total} < \tau$ and $N_{total} / S_{total} < \theta$ }{
		    \uIf{rand() $> \eta_{n}$}{
		        $P_{rand}$ $\leftarrow$ Sample in overlap area }
		    \Else{
		        $P_{rand}$ $\leftarrow$ Sample in non-overlap area }
		    $P_{nearest}$ = nearest($RRT_{nodes}$, $P_{rand}$) \\
			$P_{steer}$ = steer($P_{nearest}$, $P_{rand}$) \\
			\If{checkingObstacle($P_{nearest}$, $P_{steer}$, $M$)}{
				$RRT_{nodes}$.push\_back($P_{steer}$) \\
				$N_{total}$ = $N_{total}$ + 1 \\
				\If{explorationGain($P_{steer}$, $M$) $> MinExplorationGain$}{
				    $P_{frontier} \leftarrow P_{steer}$ }
			}
		}
	\end{algorithm}
	\vspace{-15pt}
	
    \subsection{Time Complexity Comparison with DSV Planner \cite{dsvp} }
    
    If $N$ points are successfully sampled in the non-overlap area, we suppose $k*N$ successful samples are needed in the whole sliding window. 
    We suppose a successful sampling in a sliding window requires $p$ sampling attempts.
    Therefore, the total number of required sampling attempts is $pk*N$. 
    The time cost of expanding $N$ nodes in the non-overlap area (regardless of the initialization step) can be calculated as $pk*N$ times the cost of a single while-loop in Alg. \ref{algorithm3}. 
    \vspace{-1pt}
    \begin{equation}
	\begin{aligned}
	    &T(pk*N) = T_{sample}(pk*N) + T_{nearest}(pk*N) + \\ &T_{steer}(pk*N) 
	    + T_{checkingObstacle}(pk*N) + T_{add}(k*N)
	\end{aligned}
	\end{equation}
	\vspace{-1pt}
	where $T_{sample}$, $T_{steer}$ are simple operations in linear time with a complexity of $O(pk*N)$. 
	According to \cite{choset2005principles}, the complexity of finding the nearest vertex in a kd-tree and adding a vertex to the tree is $O(n^{1-\frac{1}{d}})$ and $O(\log n)$, respectively, where $d$ represents the search dimension. 
	Therefore, the complexity of $T_{nearest}$ and $T_{add}$ is about $O((pk*N)^\frac{3}{2})$ and $O((k*N)\log (k*N))$, respectively. $T_{add}$ only needs to be executed $k*N$ times because the remaining unsuccessful sampling attempts are discarded in the ``checkingObstacle'' step. 
	$T_{checkingObstacle}$ is related to the ``steer'' step, which takes constant time. Thus, the complexity of $T_{checkingObstacle}$ is $O(pk*N)$.
    In addition, the preprocessing of the LiDAR point cloud takes constant time and can be easily integrated into the point cloud processing module, and is ignored. 
    Therefore, the overall complexity of our approach is
    \vspace{-5pt}
	\begin{equation}
	\begin{aligned}
	    &O(pk*N) + O((pk*N)^\frac{3}{2}) + O(pk*N) + O(pk*N) + \\ &O((k*N)\log (k*N)) 
	    \approx (pk)^\frac{3}{2}*O(N^\frac{3}{2}) + 3pk*O(N) + \\ &(k*N)*O(\log (k*N)).
	\end{aligned}
	\end{equation}
	\vspace{-7pt}
	
	Since $N$ is not extremely large, the first-order items of $N$ are not neglectable, we thus keep them. 
	Note that the complexity of the DSV Planner \cite{dsvp} can also represented as above. But $p$ and $k$ are different in our method and the DSV Planner.
	Generally, the successful sampling rate is proportional to the ratio of the free space to the total space in the sliding window. 
    From Section III.A, it is known that the sliding windows of both our method and the DSV Planner cover the same free space (the space that has been observed by sensors and is not occupied by obstacles). 
    Let the sliding window area of our method and the DSV Planner is  $S_{total}$ and $S_{dsv}$, respectively, 
    then the ratio of the successful sampling rate of our method to that of the DSV Planner \cite{dsvp} is equal to $S_{dsv}$/$S_{total}$. 
    When $N$ points are successfully sampled in the non-overlapping area, our method needs to sample $\frac{N}{\eta_{n}}$ points successfully in the whole sliding window, while DSV Planner \cite{dsvp} needs to sample $0.2*N+0.8*N*\frac{S_{total}}{S_{n}}$ points, where $0.2$ and $0.8$ is the bias sampling parameter of DSV Planner \cite{dsvp}. 
    
	Based on the analysis above, the $p$ value is significantly larger in the DSV Planner \cite{dsvp} than the one in our method, especially in narrow passageways. In wide open regions with obstacles, our method is still more favorable with quite much improved efficiency. In the worst case where there is no obstacle around, the adaptive sliding window method is equivalent to the fixed-sized sliding window scheme. 
    For $k$, we mainly save the calculation of sampling in overlap and redundant regions. 
    Our method almost never performs sampling in overlap or redundant area that beyond the current LiDAR scan range, as shown in Fig. \ref{firstpic}(a). 
    
    \section{Experiments}
    
    
    We obtain our evaluation results using the default settings of the Autonomous Exploration Development Environment \cite{cmuexloration}, which are the same as in the DSV Planner \cite{dsvp}. We compare our method with DSV Planner \cite{dsvp} in three scenarios: the indoor corridors, forest, and multi-storage garage scenes \footnote{A representative exploration result of the indoor corridors environment can be found at: https://youtu.be/jEY3mSrxRoU}. 
    Our algorithm's run-time is evaluated based on a 2.6GHz i7-9750H CPU. 
    We evaluate the average exploration efficiency $\varepsilon$, the average exploration volume $V$, the average exploration distance $L$, the average sampling attempts in each sliding window, the average successful samples (the samples successfully added to the RRT tree) in each sliding window, and the average sliding window duration of two methods in Table. \ref{table1}. 
    
    \begin{figure*}[tbp]
		\centering
		\subfigure[Indoor Corridors Environment]{
			\includegraphics[width=0.31\linewidth]{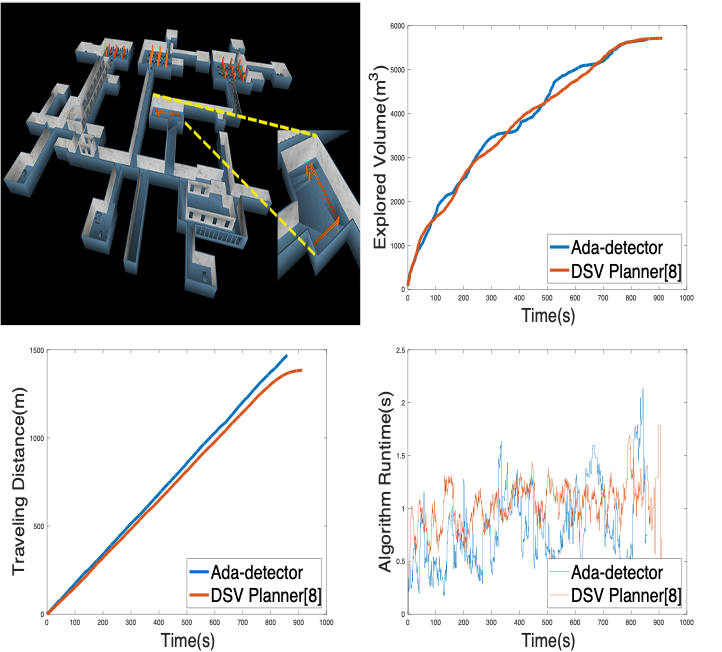}
		}
		\subfigure[Forest Environment]{
			\includegraphics[width=0.31\linewidth]{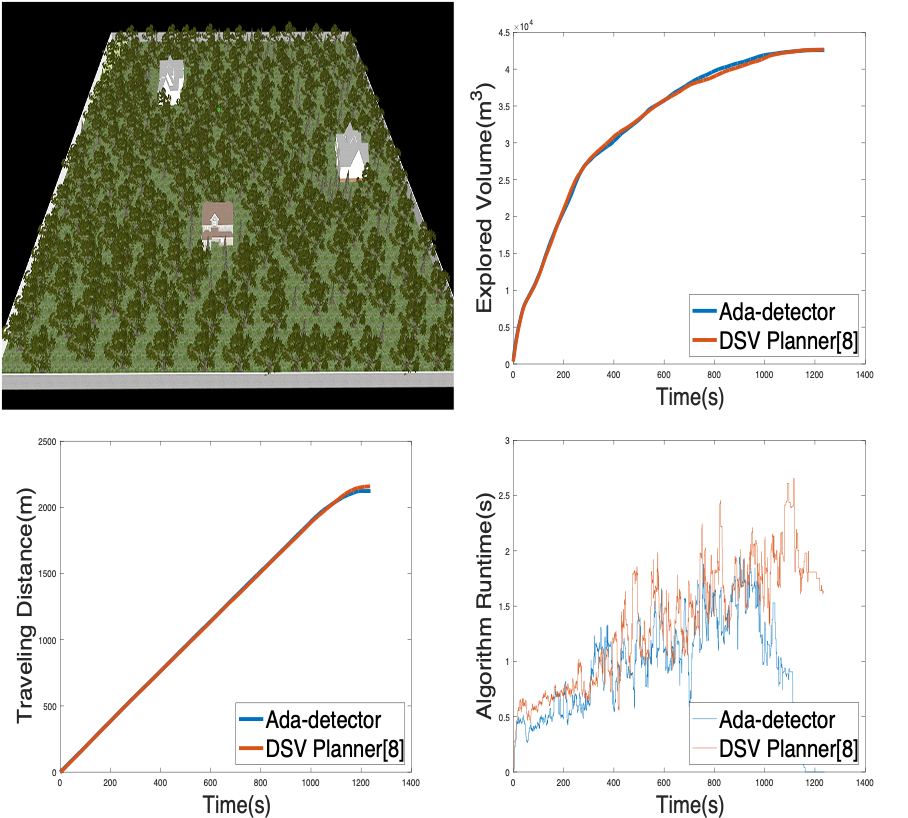}
		}
		\subfigure[Multi-storage Garage Environment]{
			\includegraphics[width=0.31\linewidth]{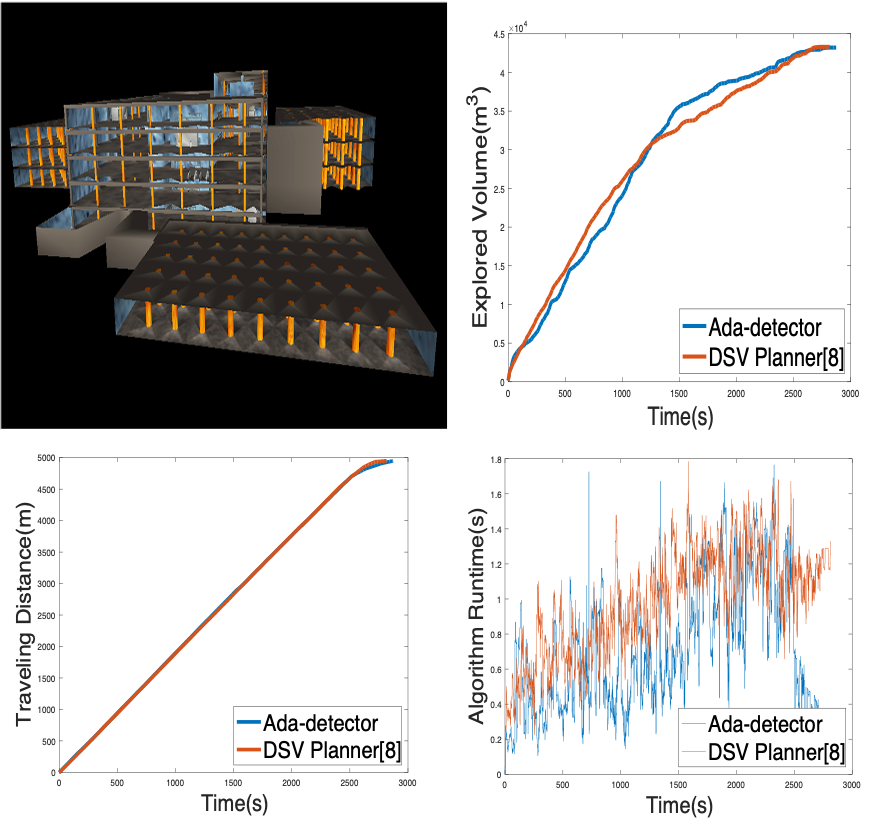}
		}
		\caption{In (a), \textbf{upper left}: the view of the indoor corridor environment. 
		\textbf{upper right}: the chart of instant exploration efficiency of two methods, where 
		the vertical axis is the exploration space volume and the horizontal axis is the exploration time. 
		\textbf{lower left}: the chart of instant exploration distance of two methods, where 
		the vertical axis is the traveling distance and the horizontal axis is the exploration time.
		\textbf{lower right}: the chart of instant algorithm runtime (including the frontier detection module and the path planning module) of two methods, where 
		the vertical axis is total running time and the horizontal axis is the exploration time. 
		(b) and (c) show the results in the forest and multi-storage garage scenarios, respectively.
		}
		\label{efficient}
	\end{figure*}
    
    \begin{figure*}[tbp]
		\centering
		\subfigure[Indoor Corridors Environment]{
			\includegraphics[width=0.31\linewidth]{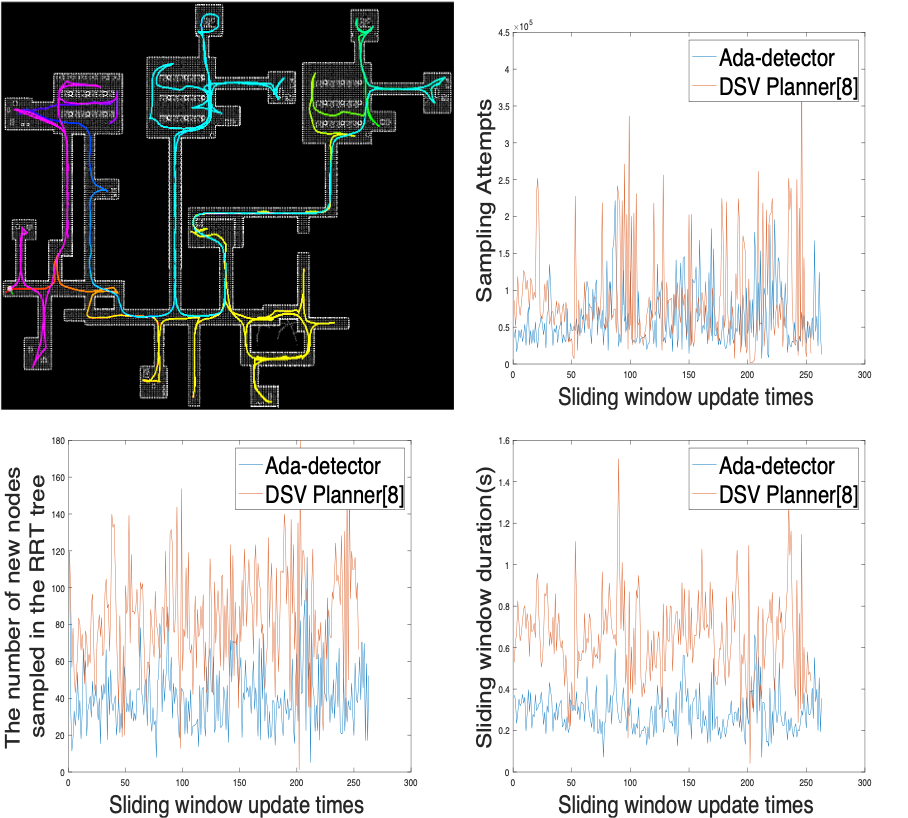}
		}
		\subfigure[Forest Environment]{
			\includegraphics[width=0.31\linewidth]{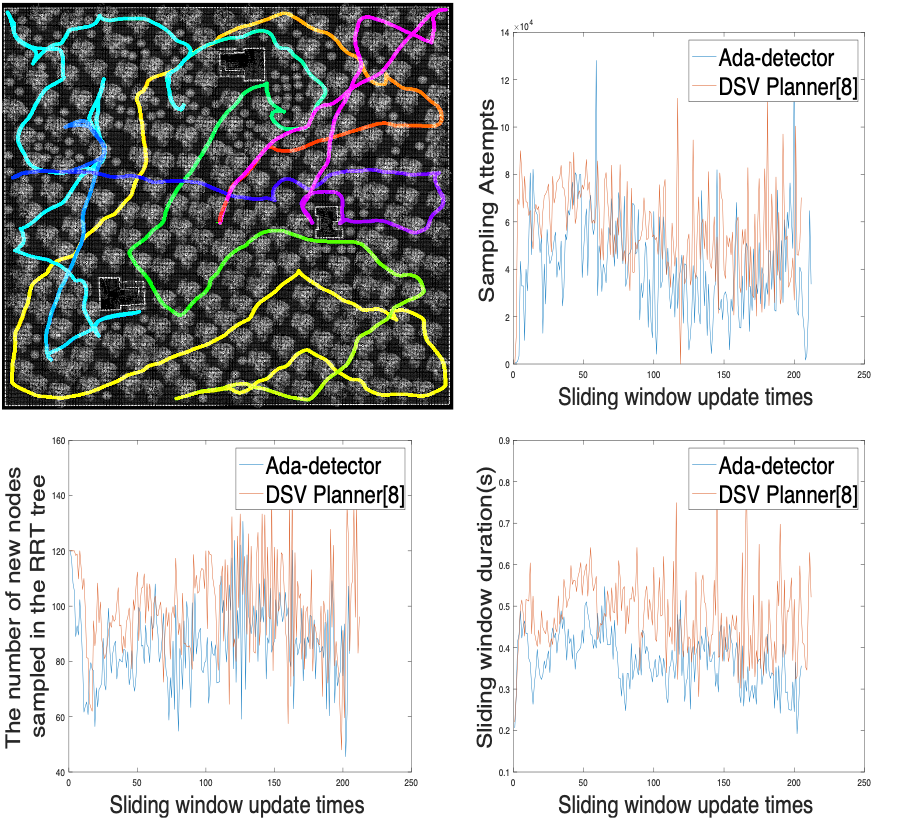}
		}
		\subfigure[Multi-storage Garage Environment]{
			\includegraphics[width=0.31\linewidth]{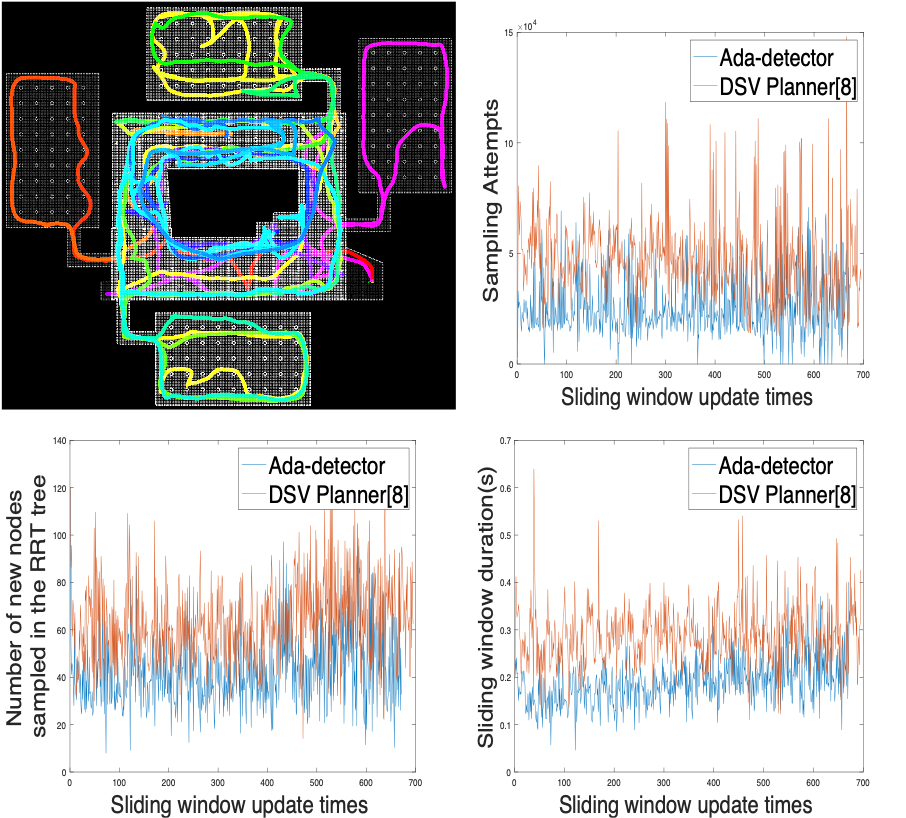}
		}
		\caption{In (a), \textbf{upper left}: the best exploration trajectory of our method in the indoor corridors simulated environments. 
		\textbf{upper right}: the chart of the average sampling attempts of two methods, where 
		the vertical axis is the number of sampling attempts in each sliding window and the horizontal axis is the sliding window update times. 
		\textbf{lower left}: the chart of the average successful samples of two methods, where 
		the vertical axis is the number of successful samples added into the RRT tree in each sliding window and the horizontal axis is the sliding window update times. 
		\textbf{lower right}: the chart of the average sliding window duration of two methods, where 
		the vertical axis is the time of sliding window duration and the horizontal axis is the sliding window update times. 
		(b) and (c) show the results in the forest and multi-storage garage scenarios, respectively. 
		}
		\label{total}
	\end{figure*}
    
    \newcommand{\tabincell}[2]{\begin{tabular}{@{}#1@{}}#2\end{tabular}}
    \begin{table*}[h!]
		\centering
        \caption{The efficiency of different methods in the simulated benchmark environments. }
        \setlength{\tabcolsep}{2mm}{
        \begin{tabular}{|l|c|c|c|c|c|c|c|c|c|c|c|c|}\hline
        &\multicolumn{6}{c|}{DSV Planner \cite{dsvp}}&\multicolumn{6}{c|}{Ada-detector}\\\hline
        &$\varepsilon$&$V$&$L$&\tabincell{c}{Sampling\\Attemps}&\tabincell{c}{Successful\\Samples}&\tabincell{c}{Sliding window\\duration}&$\varepsilon$&$V$&$L$&\tabincell{c}{Sampling\\Attemps}&\tabincell{c}{Successful\\Samples}&\tabincell{c}{Sliding window\\duration}\\
        
        Indoor&6.8&5711&1384&89062&83&0.64&7.0&5703&1380&61943&41&\textbf{0.29}\\ 
        Forest&37.7&43647&2159&56798&98&0.47&37.7&42571&2124&41123&86&\textbf{0.37} \\ 
        Garage&16.4&43311&4952&51558&63&0.31&16.3&43216&4946&25212&43&\textbf{0.18} \\ \hline
		\end{tabular} \\
	\label{table1}}
	\end{table*}
	
    The vehicle navigates at $2m/s$. Like the DSV Planner \cite{dsvp}, we run our methods 10 times. 
    The average exploration efficiency $\varepsilon (m^3/s)$ = mean of (total explored volume of a run)/(total time of a run) of 10 runs. 
    The average exploration volume $V$ and distance $L$ are the mean of the final exploration volume and distance of 10 runs, respectively. 
    The exploration data $\varepsilon, V$ and $L$ of DSV Planner \cite{dsvp} are provided by the author \cite{cmuexloration}. 
    The average sampling attempts, the average successful samples, and the average sliding window duration in each sliding window of the two methods are the mean values of 10 runs in each sliding window, respectively. 
    We draw them and the best exploration trajectories of three scenes in Fig. \ref{total}.

    
    Table. \ref{table1} shows that the successful sampling rate of our method in each sliding window is $73\%$ (indoor), $116\%$ (forest), and $139\%$ (garage) of that of the DSV Planner \cite{dsvp}, respectively. 
    In the indoor environment, due to dense obstacles, the fixed-sized sliding windows of DSV Planner \cite{dsvp} often cover the free space of other rooms (Fig. \ref{total}(a)), i.e., the free space covered by the fixed-sized sliding window of DSV Planner \cite{dsvp} is larger than that covered by the adaptive sliding window. Although higher in this case, these samples are not helpful for frontier detection because they are in other rooms, still reducing the frontier detection efficiency. 
    In the forest and garage environments, our method adaptively adjusts the size of the sliding window according to the surrounding environment structure to obtain a higher successful sampling rate. 
    
    The average number of RRT nodes added in each adaptive sliding window of three environments is $49.4\%$ (indoor), $87.8\%$ (forest), and $68.3$ (garage) of that of the DSV Planner \cite{dsvp}, respectively. 
    In three different types of scenarios, our exploration efficiency is very close to that of DSV Planner \cite{dsvp}, as shown in Fig. \ref{efficient}. 
    It indicates that both our method and the DSVP method can adequately fill the respective sliding windows. We believe that for a fully filled sliding window, both methods using the same exploration gain reach the upper limit of exploration efficiency. 
    
    Higher successful sampling rate and fewer added RRT nodes lead to higher frontier detection efficiency. 
    The average duration of each sliding window is $0.29$, $0.37$ and $0.18$ seconds, which is $45.3\%$, $78.7\%$ and $58.1\%$ of that of DSV Planner \cite{dsvp} in the indoor, forest, and garage environments, respectively.  
    We reduce the frontier detection runtime by about $40\%$ compared with the DSV Planner \cite{dsvp}. 
    
    \section{Conclusion}
    
    In this paper, we propose an adaptive RRT-based frontier detector for autonomous exploration. Our main contributions include an adaptive sliding window of RRT to improve successful sampling rate and a non-uniform sampling strategy to solve the over-sampling problem of RRT between two adjacent sliding windows.  
    We validated our method in three simulated benchmark scenarios. The experimental comparison shows that we reduce the frontier detection runtime by about $40\%$ compared with the SOTA method.

    \bibliographystyle{unsrt}
    \bibliography{bibfile}

\end{document}